# Gearbox Fault Detection through PSO Exact Wavelet Analysis and SVM Classifier

Amir Hosein Zamanian[1], Abdolreza Ohadi[2]

[1] MSc student, Mechanical Engineering Department, Amirkabir University of Technology; zamanian.amir@gmail.com
[2] Associate Professor, Mechanical Engineering Department, Amirkabir University of Technology; a_r_ohadi@aut.ac.ir

**Abstract**
Time-frequency methods for vibration-based gearbox faults detection have been considered the most efficient method. Among these methods, continuous wavelet transform (CWT) as one of the best time-frequency method has been used for both stationary and transitory signals. Some deficiencies of CWT are problem of overlapping and distortion of signals. In this condition, a large amount of redundant information exists so that it may cause false alarm or misinterpretation of the operator. In this paper a modified method called Exact Wavelet Analysis is used to minimize the effects of overlapping and distortion in case of gearbox faults. To implement exact wavelet analysis, Particle Swarm Optimization (PSO) algorithm has been used for this purpose. This method have been implemented for the acceleration signals from 2D acceleration sensor acquired by Advantech™ PCI-1710 card from a gearbox test setup in Amirkabir University of Technology. Gearbox has been considered in both healthy and chipped tooth gears conditions. Kernelized Support Vector Machine (SVM) with radial basis functions has used the extracted features from exact wavelet analysis for classification. The efficiency of this classifier is then evaluated with the other signals acquired from the setup test. The results show that in comparison of CWT, PSO Exact Wavelet Transform has better ability in feature extraction in price of more computational effort. In addition, PSO exact wavelet has better speed comparing to Genetic Algorithm (GA) exact wavelet in condition of equal population because of factoring mutation and crossover in PSO algorithm. SVM classifier with the extracted features in gearbox shows very good results and its ability has been proved.

**Introduction**
The existence of fault in rotating systems can make hazardous damage for operators or systems. So, early detection of fault in the systems may prevent casualties or stopping systems. One of the common tools in industry is gearboxes, which may contain faults. There have been many investigations by researchers on case of gear fault detection.
There are several methods in fault detection, in one view the signals categorized to vibrating signals, acoustic emitted (AE) signals and electric current signals obtained from systems [1]. In another view the method of signal processing categorized to time domain, frequency domain or time-frequency domain analysis. Although features extracted from time domain analysis (i.e. statistical features) can detect fault (same as [1-3]), but generally these methods need extensive data mining procedure for feature selection so different types of data mining algorithm and decision trees are applied with these features. On the other hand, there exist some problems for frequency methods (e.g. Fast Fourier Transform). Indeed, frequency methods eliminate the effect of time in the signal, so these methods only indicate the frequencies and its intensity, but they do not reveal how many frequencies exist and with what intensity these frequencies repeat. These methods are mainly applied to detect for harmonics in signals.
For the sake of overcome these disadvantages, the time-frequency methods (Gabor transform, Hilbert-Haung transform, Wavelet transform, etc) take lots of attractions for researchers who would like to study signal in frequency content but they wouldn't like to eliminate time information. Generally, time-frequency methods provide energy distribution of signals; hence, these methods seem to be what researchers are looking for.
Among these, continuous wavelet transform (CWT) is mainly considered as an effective tool for vibration-based signal processing for fault detection. CWT provides a multi-resolution in time-frequency analysis for characterizing the transitory features of non-stationary signals [4]. Two problems, "distortion" and "overlapping", suffer CWT which are completely discussed by Tse et al. [4].
Tse et al. [4], to solve the problem of distortion and overlapping, firstly used the term of "exact wavelet analysis". They provided an opportunity to find most appropriate scale and wavelet daughter shape in each time frame of inspected signal by a GA-based optimization process by defining an objective function which indicates similarity between wavelet coefficient and the inspected signal. So, for any time frame, GA is allowed to find the scale parameter and some parameters related to the shape of the wavelet presented by Brode [5].
The authors of this paper believe that in best condition, when the parameters is found by GA, the exact wavelet cannot find the most appropriate wavelet although it can find the best wavelet family presented by Brode [5]. So, here we use Morlet wavelet and reduce the optimization parameters to one parameter; scale variable. In this case, the optimization space reduces from $\Re^3$ to $\Re^1$. This speeds up the optimization process and decreases the probability of local optima. To speed up process even more, the GA has been replaced by PSO algorithm.

Support vector machine (SVM) has been used in many applications of machine learning because of high accuracy and good generalization capability. SVM classifies better than artificial neural networks (ANN) because of the principle of risk minimization. In ANN, traditional empirical risk minimization (ERM) is used on training data set to minimize the error, whereas in SVM, structural risk minimization (SRM) is used to minimize an upper bound on the expected risk [3].

SVM has been successfully applied in automated detection of machines [1,2] based on learning patterns. Jack and Nandi [2] compared SVM by ANN for classification of fault in bearing by feature (statistical and spectral) selection based on genetic algorithm (GA). They showed that ANN tends to be faster to train and more robust than SVM in case of bearing vibration signals. In contrary, Samanta [1] made an investigation based on statistical features acquired form gears, the effectiveness of both ANN and SVM was compared. It has been shown that, for most of the cases considered, the classification accuracy of SVM is better than ANN when the GA-based feature selection has not been used. However, with GA-based feature selection the performance of both classifiers is comparable.

In part two of the current work, a brief introduction to exact wavelet transform is presented, in part three PSO is introduced, part four is reserved for SVM classifier and in part five the algorithm is applied on an experimental setup test. Summary and conclusions take final part of this paper.

**Exact Wavelet Analysis**
There exists two kind of exact wavelet analyses, the first method utilizes the concept of ''maximum matching mechanism'' to determine the most appropriate coefficients to represent the inspected raw signal. In CWTs with a given signal, within the selected time frame, if a daughter wavelet, which is generated by a particular scale, has the largest value of wavelet coefficient, it often implies that the shape of that daughter wavelet can match the shape of the inspected signal better than other daughter wavelets generated by other scales [4].

The advantage of this method is its simplicity and higher computational speed, whereas its disadvantage refers to the fact that it cannot find appropriate daughter wavelet with the geometric shape exactly similar to the inspected signal within the selected time frame. In addition, selection of mother wavelet is not adaptive to the inspected signal [4].

The second method is aimed to provide a direct measure of the similarity in shapes between the daughter wavelet and the inspected signal. Instead of using the largest value of wavelet coefficient, the ''normalized dot product'' of the daughter wavelet and the inspected signal is adopted for measuring their similarity in shape [4].

Continuous wavelet transform is defined as,

$$W_\psi(a,b) = \int_{-\infty}^{+\infty} x(t) \overline{\psi}_{a,b}(t) dt, \quad (1)$$

where

$$\psi_{a,b}(t) = |a|^{-\frac{1}{2}} \psi\left((t-b)/a\right) \quad (2)$$

is a window function called mother wavelet and $a$ and $b$ are real-valued parameters, $b$ is the translation parameter indicating the position, and $a$ is the scale parameter.

In this paper, a modified version of second method has been selected by considering that most appropriate wavelet is not selected by exact wavelet. The Morlet wavelet (Fig. 1) is considered as most appropriate wavelet because of its similarity to response of impulse function. Therefore, in each time frame only scale is optimized and consequently, in each translation, the algorithm gives the most appropriate scale. The advantage of this method is that the optimization algorithm does not waste time for calculation of wavelet parameters, so optimization process goes faster.

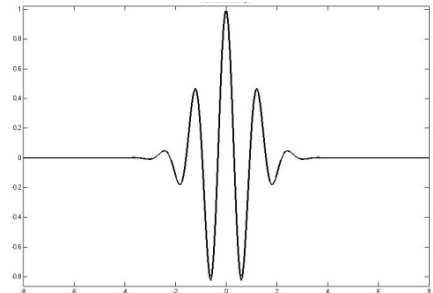
Figure 1: Morlet wavelet.

For each selected time frame, the similarity of wavelet and portion of signal is calculated by normalized dot product between the exact wavelet coefficients and the portion of inspected signal, a fitness index can be obtained to evaluate the degree of matching. The index is calculated using a cosine function of two vectors [4].

$$\cos(\mathbf{C},\mathbf{X}) = \frac{\sum_{i=1}^{n} c_i x_i}{\sqrt{\sum_{i=1}^{n} c_i^2 \sum_{i=1}^{n} x_i^2}}, \quad 1 < n < N, \quad (3)$$

where $\mathbf{C}$ and $\mathbf{X}$, stand for the vectors of the wavelet coefficients and the portion of the inspected signal, respectively. The variables $c_i$ and $x_i$ represent the elements of the vectors and $N$ is the number of signal samples.

The calculated index from the fitness function provides a measure to evaluate the similarity of the two vectors not only in their magnitudes but also in their geometrical shapes. The higher indexes of fitness function indicates that more similarity exist between the derived wavelet and the portion of the inspected signal. The index of the cosine function approaches to 1 when a perfect match exist, whilst the zero value index shows a mismatch [4].

**Particle Swarm Optimization**
Particle swarm optimization (PSO) is a population based stochastic optimization technique developed by Kennedy and Eberhart [6] in 1995, inspired by social behavior of bird flocking or fish schooling. PSO started with population of random solutions and updating the solutions in next generation to find optimal solution [7]. The potential solutions in PSO are called particles.

These particles fly thorough hyper space of the problem by following best particles [7].

All particles have fitness values, which are evaluated by the fitness function, and have velocities, which direct the flying of the particles. The particles fly through the problem space by following the current optimum particles. PSO is initialized with a group of random particles and then searches for optima by updating generations [7]. In each iteration, particles are updated by following two "best" values. The first one is the best solution (fitness) which a particle is achieved so far (**pb**) and the second one is the best global solution obtained so far in all particles of the population (**gb**) [7]. Velocity and position of particles are updated with the following equations:

$$\mathbf{v}_i^{t+1} = \mathbf{v}_i^t + c_1 r_1 \left( \mathbf{pb}_i^t - \mathbf{p}_i^t \right) + c_2 r_2 \left( \mathbf{gb}^t - \mathbf{p}_i^t \right), \quad (4)$$

$$\mathbf{p}_i^{t+1} = \mathbf{p}_i^t + \mathbf{v}_i^{t+1}, \quad (5)$$

where $\mathbf{v}_i$, $\mathbf{p}_i$ are the $i$-th particle velocity and particle position (solution), respectively. $\mathbf{pb}_i^t$ and $\mathbf{gb}^t$ are already defined, $r_1$, $r_2$ are random numbers between (0,1) and $c_1$, $c_2$ are learning factors (usually $c_1 = c_2 = 2$) [7]. Velocities of particles on each dimension are constrained to a maximum velocity $\|\mathbf{V}_{max}\|$.

**Support Vector Machine**

The foundations of support vector machines (SVM) have been developed by Vapnik in 1995 [8] and are gaining popularity due to many attractive features, and promising empirical performance. The formulation embodies the structural risk minimization (SRM) principle, which has been shown to be superior to traditional empirical risk minimization (ERM) principle, employed by conventional neural networks [9].

The concept of support vector machine is extensive, a brief introduction of SVM presented here, the readers are referred to [9, 10] for more details.

Without loss of generality, the classification problem can be restricted to consideration of the two-class problem. SVM can be considered to create a line or hyper-plane between two set of data for classification [1].

Consider the problem of separating the set of training vectors belonging to two separate classes

$$D = \left\{ \left( \mathbf{x}^1, y^1 \right), ..., \left( \mathbf{x}^l, y^l \right) \right\}, \quad x \in \mathbb{R}^n, y \in \{-1, 1\}, \quad (6)$$

with a hyper-plane,

$$\mathbf{w}.\mathbf{x} + b = 0. \quad (7)$$

In the case of two-dimensional situation, the action of the SVM can be explained easily. In this situation, SVM try to find a line which separate two classes of data (feature sets) by a line (hyper plane). This line separates data into two parts so the data on the right hand belong to one class (Class A) and the data on the left hand belong to the other class (Class B). Many lines have the ability to separate data truly. However, SVM try to find that line which has the maximum Euclidean distance between the nearest data to this line either in Class A and B. The data, which has the minimum distance to this line, are called support vectors (SVs) that are shown in Fig. 2. Since training SVM with SVs is sufficed, the rest of data can be neglected.

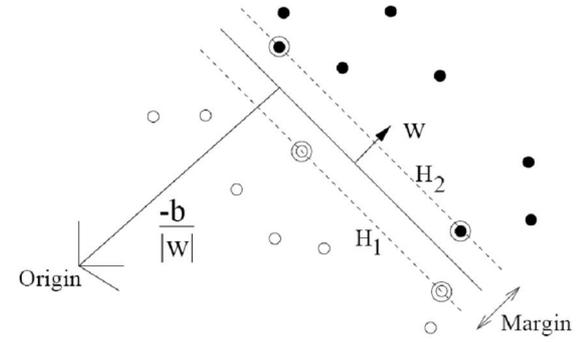

Figure 2: Linear separating plane for classification, the support vectors are circled. [10]

The SVs are located in two parallel lines, which are parallel to the separating line. The margin equations for class A and B are as follows:

$$\mathbf{w}.\mathbf{x} + b = 1 \quad \text{(Class A)}, \quad (8)$$

and

$$\mathbf{w}.\mathbf{x} + b = -1 \quad \text{(Class B)}. \quad (9)$$

Once the SVM has been trained, a decision function in Eq. (10) determines that each test sample belongs to which side of decision boundary (i.e. which class).

$$f(x) = \text{sgn}(\mathbf{w}.\mathbf{x} + b), \quad (10)$$

The SVM training is obtained by optimizing an objective function that is presented in Eq.(11)

$$L = \frac{1}{2} \|\mathbf{w}\|^2 - \sum_{i=1}^{l} \alpha_i y_i \left( \mathbf{w}.\mathbf{x}_i + b \right) + \sum_{i=1}^{l} \alpha_i, \quad (11)$$

where $l$ is the number of training sets, and $\alpha_i$ is Lagrange multipliers' coefficients obtained by the following constraints.

$$y_i \left( \mathbf{w}.\mathbf{x}_i + b \right) \geq 1, \quad (12)$$

The solution can be obtained as follows

$$\mathbf{w} = \sum_{i=1}^{l} \alpha_i y_i \mathbf{x}_i = \sum_{i=1}^{l} v_i \mathbf{x}_i, \quad (13)$$

where

$$v_i = \alpha_i y_i \quad (14)$$

Substituting of Eq. (13) to Eq.(10) leads to Eq.(15).

$$f(x) = \text{sgn}\left( \sum_{i=1}^{l} v_i (\mathbf{x}_i.\mathbf{x}) + b \right) \quad (15)$$

The set of vectors is said to be optimally separated by the hyper plane if it is separated without error and the distance between the closest vector to the hyper plane is maximal [9]. In the case of non-separable data with linear hyper plane, a hyper plane should be defined that allows linear separation in the higher dimension (corresponding to nonlinear separating hyper planes) [1]. To do this, the data should be mapped to some other spaces, using a mapping $\Phi$.

$$\Phi : \mathfrak{R}^n \rightarrow H$$

where $H$ is a Hilbert space (as a generalization of Euclidean space [10]), so by defining $\Phi(\mathbf{x})$, the data can be transformed to the new space, by defining a kernel function $K(\mathbf{x}_i, \mathbf{x})$ in Eq.(16). The former equations can be modified by changing the dot product of $\mathbf{x}_i.\mathbf{x}$ to $K(\mathbf{x}_i, \mathbf{x})$.

$$K(\mathbf{x}_i, \mathbf{x}_j) = \Phi(\mathbf{x}_i) \cdot \Phi(\mathbf{x}_j) \qquad (16)$$

The defined kernel function emphasize that $\Phi(\mathbf{x})$ is not necessary to be known explicitly. Therefore, Eq.(15) changes to Eq.(17).

$$f(x) = \mathrm{sgn}\left(\sum_{i=1}^{l} v_i K(\mathbf{x}_i, \mathbf{x}) + b\right) \qquad (17)$$

There exists different kernel functions, a common function called Radial Basis Function (RBF) which is used in this work, given by Eq.(18).

$$K(\mathbf{x}_i, \mathbf{x}_j) = \exp\left(-\|\mathbf{x}_i - \mathbf{x}_j\|^2 / 2\sigma^2\right) \qquad (18)$$

The parameter $\sigma$ is width of RBF kernel, which is an important parameter in classification performance and can be determined by statistical computations or by iterative process to choose optimum value [1].

For the non-separable data, where overlap exists between the classes, the range of parameters $v_i$ should be bounded to reduce the effect of outliers on the boundary defined by SVs (i.e. $|v_i| < C$).

For separable cases, $C$ is infinity while for non-separable cases, it may be varied, depending on the number of allowable errors in the trained solution: high $C$ permits few errors while low $C$ allows a higher proportion of errors in the solution [1,10].

**Experimental Results**

To evaluate the efficiency of exact wavelet, the algorithm was implemented on an experimental vibration signal of gearbox. This process was done for two conditions, normal and chipped tooth gear (as shown in Fig. 3). In the chipped tooth case, 50% of a tooth profile from top to pitch circle (addendum) was eliminated with linear slop from top to pitch circle.

The signals were acquired from a gearbox setup test designed in Amirkabir University of Technology (Tehran Polytechnic) for this proposes, as shown in Fig. 5. The vibrating signals were obtained from 2D accelerometer (ADXL210JQC) mounted on gearbox frame (Fig. 4). Sampling frequency was set as 10kHz. The acceleration frequency content is in the range of 0~5kHz. These obtained signals were fed to A/D converter (Advantech™ PCI-1710, 12-bit, 100kS/s) and was recorded by real-time workshop of MATLAB software. The gearbox rotates with nominal speed of 3-phase electromotor (1420 RPM). The driver ($N_1 = 15$) and driven ($N_2 = 110$) gear provide speed ratio of 7.33:1 for gearbox. The disk brake system has been considered to provide appropriate load on the system.

Although number of samples for each feature sets is completely arbitrary, however it is not appropriate to select small portion of signal. So each feature sets considered here contains 1250 signal samples (approximately equal to 3 round of driver gear). 80 samples for each normal and chipped tooth gear condition (totally 160 feature sets) have been created.

The exact wavelet was implemented by PSO algorithm with parameters shown in Table 1. Range of scale is considered between 1 and 32. To create a feature set, distribution of scales in 1250 samples was counted and divided to 16 ranges of scales. Fig. 6 shows two feature sets, one of them corresponds to normal gear and another one belongs to chipped tooth gear. The difference of feature sets is obviously apparent.

Distributions of number of data point belong to each scale level for normal and chipped tooth gear conditions create 16 feature. These feature sets have been used to train SVM classifier. The programming of these procedures was done by MATLAB.

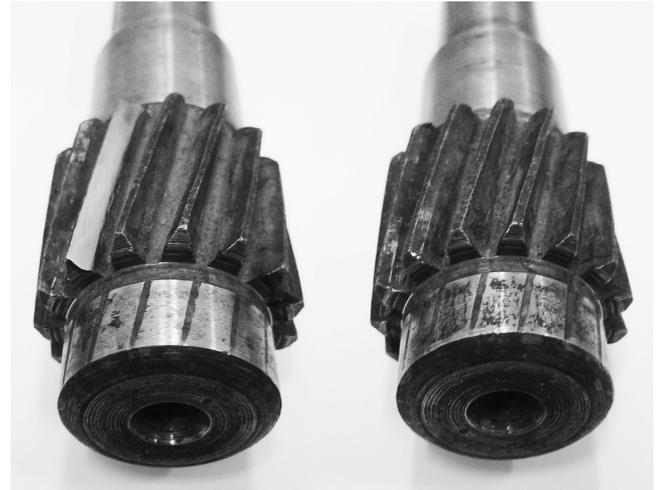

a) Chipped tooth gear      b) Normal gear.
Figure 3: Tested gears.

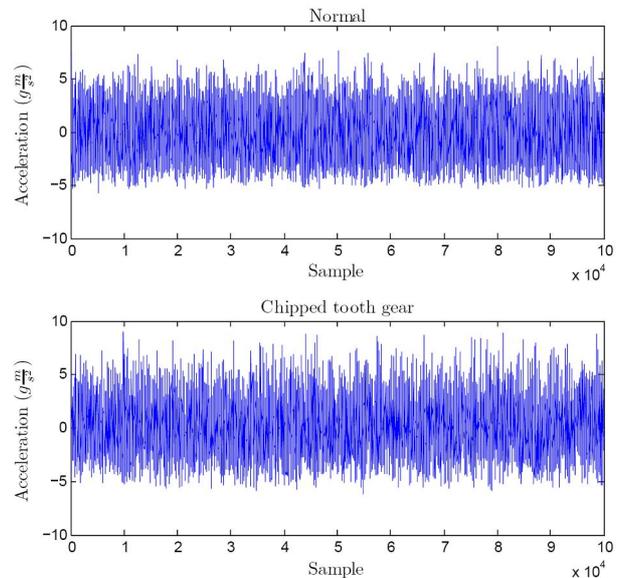

Figure 4: The accelaration signal acquried from gear box for Normal and chipped tooth gear condition.

Table 1: PSO parameters

| $c_1$ | $c_2$ | $\|\mathbf{V}_{max}\|$ | Space | Population (bees) | Maximum generation (flights) | Goal | Stall time limit | Time limit |
|---|---|---|---|---|---|---|---|---|
| 2 | 2 | $\infty$ | 1 to 32 | 20 | 50 | $\infty$ | 20 sec | 30 sec |

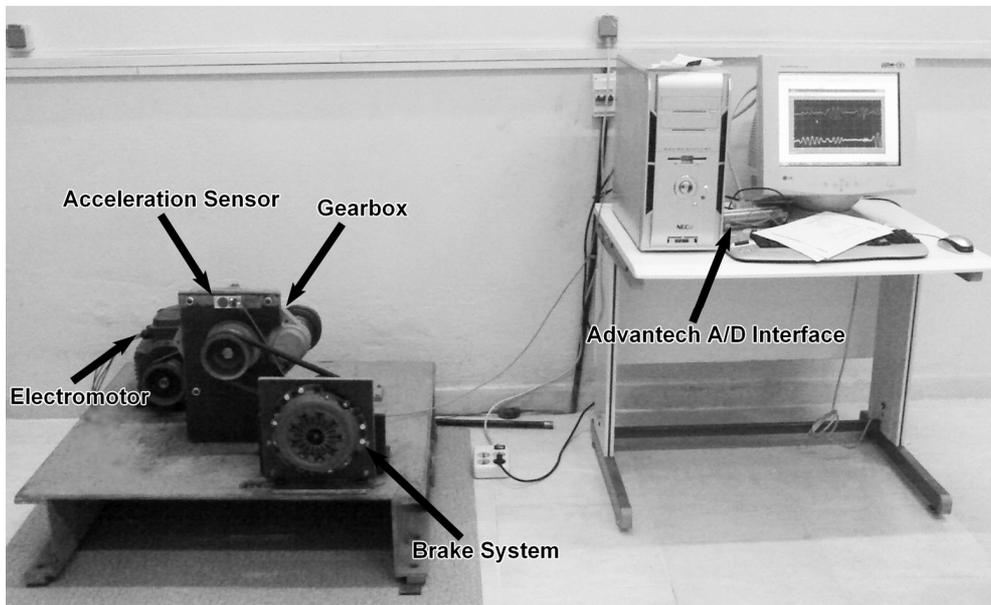

Figure 5: The gearbox experimental setup test (Acoustics Research Lab., Mechanical Engineering Department, Amirkabir University of Technology).

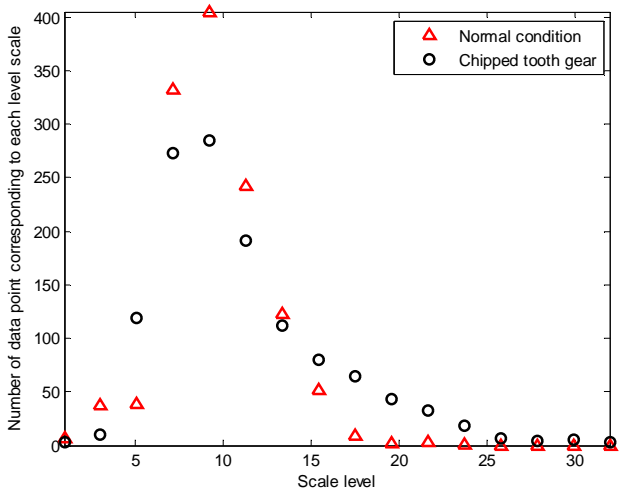

Figure 6: Distribution of number of data point belong to each scale level for normal and chipped tooth gear condition

SVM classifier was trained with 60 feature sets, and 100 feature sets were remained for test success measurement. The result of SVM classifier is presented in Table 2.

Table 2: Performance of SVM (16 feature sets)

| SVM (RBF) $\sigma = 0.5$ | | SVM (RBF) $\sigma = 1.0$ | | SVM (RBF) $\sigma = 1.5$ | | SVM (Linear) | |
|---|---|---|---|---|---|---|---|
| Training success (%) | Test success (%) | Training success (%) | Test success (%) | Training success (%) | Test success (%) | Training success (%) | Test success (%) |
| 100 | 94 | 100 | 99 | 100 | 100 | 100 | 100 |

It is clear in Table 2 that the feature sets is completely linear separable, so use of linear SVM bring 100% success in classification. Nonlinear SVM with RBF bring excellent classification when the $\sigma$ parameter increases, because SVM tends to have a linear manner.

**Comparison of PSO and GA exact**
The authors applied the exact wavelet by GA, with similar parameter of PSO and the computational time was almost 40 times more in case of GA exact wavelet analysis. However, GA solutions had better optimization performance in finding most appropriate scale and PSO generally filled in local maxima. It should be noted that the PSO and GA solutions were close to each other and the difference in solutions does not affect the mentioned feature extraction. The solution for 20 sample of signal is plotted in Fig. 7. Because of stochastic nature of both GA and PSO methods, the solutions shown in Fig. 7 might change in different runs.

Table 3: GA parameters

| Population | Elite | Mutation | Crossover | Generation | Function's Tolerance | Space (initial range) |
|---|---|---|---|---|---|---|
| 20 | 4 | Uniform (Probability 1%) | Scattered (80% Fraction) | 50 | 1e-9 | 1 to 32 |

**Conclusions**
In this paper, a modified version of exact wavelet analysis, which speeds up the exact wavelet process, is introduced. It has been shown that PSO exact wavelet can speed up the process almost 40 times comparing to GA. It is also shown than parameters of exact wavelet can be reduced by using of appropriate wavelet (i.e. Morlet wavelet) to prevent designing wavelet. That speed up more, without destroying feature sets.
The SVM classifier has been used with 16 feature, although SVM with radial bases can classify with test success of 100% by considering suitable sigma as shown in table 1. However, it has been shown that the feature sets are linear separable, so linear SVM in case of classification of normal and chipped tooth gear will suffice. The main problem that suffers exact wavelet even with this modified form is computational time that cannot be used in real time application.

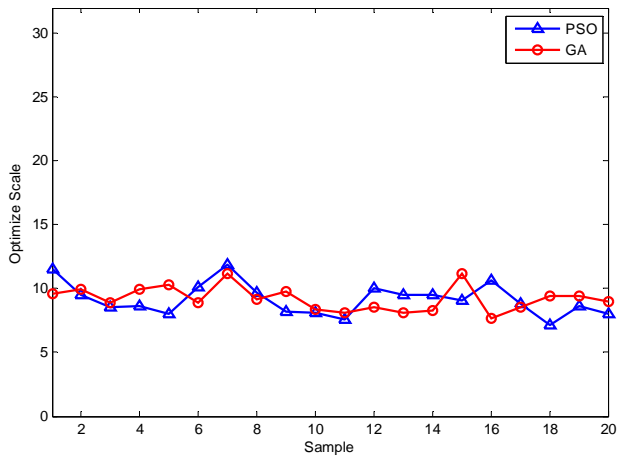

Figure 7: Comparison of PSO and GA optimization solution in exact wavelet analysis for 20 sample of acceleration signal.

**Acknowledgment**
The authors should appreciate to Mr. Mohammad Sarikhani, educational instructor in Amirkabir University of Technology's Vibration and Dynamics of Machine Lab. for his valuables consultations and Mr. Mohammad Rostami Ghomi, MSc electronic student of Sharif University of Technology for designing some electronics equipment for this project.